\documentclass[11pt,a4paper]{article}
\usepackage{authblk}
\usepackage[hyperref]{acl2017}
\usepackage{times}
\usepackage{latexsym}
\usepackage[utf8]{inputenc}
\usepackage[T1]{fontenc}
\usepackage{graphicx}
\usepackage{wrapfig}
\usepackage{multicol}
\usepackage{tabularx}

\usepackage{url}

\aclfinalcopy 
\def\aclpaperid{173} 

\title{Augmenting a BiLSTM tagger with a Morphological Lexicon\\and a Lexical Category Identification Step}

\author[1,2]{Steinþór Steingrímsson}
\author[1]{Örvar Kárason}

\author[1]{Hrafn Loftsson}
\affil[1]{Department of Computer Science, Reykjavik University,
Iceland}
\affil[2]{The Árni Magnússon Institute for Icelandic Studies, Reykjavik, Iceland}
\affil[ ]{\texttt {\{steinthor18,orvark13,hrafn\}@ru.is}}

\date{July 18th 2019}

\begin{document}
\maketitle
\begin{abstract}
  Previous work on using BiLSTM models for PoS tagging has primarily focused on small tagsets. We evaluate BiLSTM models for tagging Icelandic, a morphologically rich language, using a relatively large tagset. Our baseline BiLSTM model achieves higher accuracy than any previously published tagger not taking advantage of a morphological lexicon. When we extend the model by incorporating such data, we outperform previous state-of-the-art results by a significant margin.
  We also report on work in progress that attempts to address the problem of data sparsity inherent in morphologically detailed, fine-grained tagsets. We experiment with training a separate model on only the lexical category and using the coarse-grained output tag as an input for the main model. 
  This method further increases the accuracy and reduces the tagging errors by 21.3\% compared to previous state-of-the-art results.
  Finally, we train and test our tagger on a new gold standard for Icelandic. 
\end{abstract}

\section{Introduction}

Bidirectional long short-term memory (BiLSTM) models have in recent years been shown to be effective for various sequential labelling tasks, including Part-of-Speech (PoS) tagging \cite{DBLP:journals/corr/LingLMAADBT15,DBLP:journals/corr/PlankSG16}.

BiLSTMs are an extension of general LSTMs \citep{Hochreiter:1997:LSM:1246443.1246450} that perform better on sequences where the complete input sequence is available. Two LSTMs are trained on the input sequence, one on its natural reading order and the other on its reverse \citep{Graves:2005:SIF:1120509.1120527}. 
 In addition to word embeddings (WE), character embeddings were first used for tagging with BiLSTMs by \citet{DosSantos:2014:LCR:3044805.3045095}. This entails not only examining the sequence of words in a sentence during training but also the sequences of characters within those words.
 
 In this paper we use BiLSTM models, with both word and character embeddings, to train a PoS tagger for a morphologically rich language, Icelandic, with a fine-grained tagset of 565 morphosyntactic tags. Only a small portion of previous work using neural networks for PoS tagging has focused on languages with rich morphology and large tagsets, e.g. \citet{W17-6304}.

Various taggers have been developed for Icelandic: data-driven taggers \citep{helgadottir2005}, a rule-based tagger (IceTagger)   \citep{loftsson_2008}, and a hybrid tagger \citep{Loftsson2009ImprovingTP}. Prior to the work presented here, an averaged perceptron tagger, IceStagger \citep{Loftsson2013tagginga}, was the current state-of-the-art tagger, achieving an accuracy of 93.84\% by employing a morphological lexicon and external word embeddings. 

This paper presents the first deep neural network tagger for Icelandic. We evaluate three models. First, we confirm the effectiveness of a BiLSTM model for PoS tagging using a fine-grained tagset. Second, we supplement the base model with an external morphological lexicon, thereby obtaining state-of-the-art results. Third, we propose an approach to further increase the accuracy by creating a coarse-grained tagset from the fine-grained one and using the resulting tagset to devise a two-step process. This approach is to our best knowledge novel in the context of neural network tagging. 
Specifically, we train a separate model on only the lexical category and use the coarse-grained output tag as an input into the main model. Combined, this results in an overall tagging accuracy of 95.15\%, which is equivalent to an error reduction of 21.3\% compared to the previous state-of-the-art. 
Finally, we train and test our model on a new gold standard for Icelandic, MIM-GOLD. The new standard is larger than the older one, IFD (see Section \ref{sec:data}), and contains more diverse texts. We achieve an accuracy of 94.17\% on MIM-GOLD. 

\section{Data}
\label{sec:data}

In this section, we describe the data and the tagset used in our work.

\textbf{The IFD Corpus:} The taggers developed for Icelandic so far have all been trained and tested on the Icelandic Frequency Dictionary (IFD) corpus \citep{magnusson1991islensk}, a balanced corpus containing about 590 thousand tokens. The IFD corpus was collected in the early 1990s and contains texts from published books, primarily fiction (60\%) but also biographies (20\%) and scholarly work (20\%). As with the other taggers referenced in this paper, we use the so-called \textit{corrected version} of the corpus, with the reduced tagset (565 tags) and ten-fold split from \citet{Loftsson2009ImprovingTP}.\footnote{IFD can be downloaded from \url{http://malfong.is/?pg=ordtidnibok}.}
The morphosyntactic tags in this tagset are mnemonic encodings, i.e. character strings where each character has a particular function.
The first character denotes the \emph{lexical category}.
For each category there is a predefined number of additional characters (at most six), which describe morphological features, like \emph{gender}, \emph{number} and \emph{case} for nouns; \emph{degree} and \emph{declension} for adjectives; \emph{voice}, \emph{mood} and \emph{tense} for verbs, etc.
To illustrate, consider the word form \emph{maður} ``man''.
The corresponding tag is \emph{nken}, denoting noun (\emph{n}), masculine (\emph{k}), singular (\emph{e}), and nominative (\emph{n}) case.

\textbf{The MIM-GOLD corpus:} MIM-GOLD\footnote{MIM-GOLD can be downloaded from \url{http://malfong.is/?pg=gull}.} \citep{Loftsson2010DevelopingAP}, a subset of the MIM corpus \citep{Helgadttir2012TheTI}, contains a greater diversity of texts than the IFD corpus. In addition to texts from published books, it contains texts from news media, blogs, parliamentary speeches and more. Furthermore, MIM-GOLD contains approximately 1 million running words, about twice as many as IFD. The tagset used in MIM-GOLD consists of the same reduced tagset of 565 tags, mentioned above.

\textbf{Morphological Lexicon:} The Database of Modern Icelandic Inflections (DMII) is a lexicon of about 280 thousand paradigms and close to six million inflectional forms \citep{Bjarnadottir2012}. The output from the database used in this project contains word form and morphological features. By incorporating DMII, the average unknown word rate in testing, using the IFD ten-fold split, drops from 6.8\% to 1.1\% \cite{DBLP:conf/ranlp/LoftssonHR11}.

\section{The Three Models}
\subsection{Word and Character Embeddings}
\label{sec:wordCharEmbeddings}
Word embeddings are vector representations of words based on their context in training data.
Adding recurrent character embeddings has been shown to significantly improve performance for handling of unknown words (e.g. \citealt{DBLP:journals/corr/PlankSG16, DosSantos:2014:LCR:3044805.3045095}). For each word, both forward and backward expressions are generated, containing the sequence of characters in the word, as well as word initial and word final markers. This helps the model grasp morphological details.

In our baseline model, which is similar to \citet{DBLP:journals/corr/PlankSG16}, both word embeddings and recurrent character embeddings are used as input. The character embeddings for a given word are input into a BiLSTM. The output from the BiLSTM is concatenated to the word embedding and the combined vector input into another BiLSTM, whose output is input into a hidden layer. The hidden layer feeds the output layer, which selects a PoS tag.



\subsection{Using Data from an External Morphological Lexicon}
\label{sec:DMII}

\citet{D17-1076} replicated \citet{DBLP:journals/corr/PlankSG16} using a collection of corpora annotated with fine-grained tagsets of varying sizes, in contrast to the coarse-grained Universal Dependencies (UD) tagset in the previous study (17 tags). The replication confirmed the superior performance of the BiLSTM tagger, also on fine-grained tagsets. Furthermore, they found that the advantages of the BiLSTM tagger over other taggers grow proportionally with the tagset size of the corpus. However, they also claim that for large tagsets of morphologically rich languages, hand-crafted morphological lexicons are still necessary to reach state-of-the-art performance.

Using a morphological lexicon has become common practice for enriching training data for PoS taggers. \citet{biblio:HaMorphologicalTagging2000} marked the importance of this for morphologically rich languages.
It was first done for Icelandic in \citet{DBLP:conf/ranlp/LoftssonHR11}. 

\begin{figure}[t]
\centering
\includegraphics[width=0.43\textwidth]{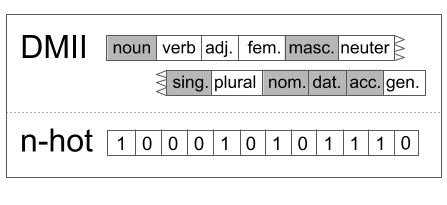}
\caption{\label{fig-n-hot}A partial n-hot vector and the corresponding features from DMII. The example shows 12 features, including the active features for the word form \emph{strætó} ``bus''. The word, a noun, has the same form for nominative, dative and accusative and therefore all corresponding labels are activated. An actual vector in our model has 61 labels, which are either active, \texttt{1}, or inactive, \texttt{0}.}
\end{figure}

\citet{W17-6304} first used morphological lexicons as supplemental input for PoS tagging with BiLSTM taggers and showed that it yields consistent improvement. Following their work, we extend the baseline model by adding an input layer that contains token-wise features obtained from the DMII lexicon (see Section \ref{sec:data}). The input vector for a given word is an n-hot vector where each active value corresponds to one of 61 possible labels in the lexicon. An example of an n-hot vector is given in Figure \ref{fig-n-hot}.

The vector is concatenated to the two vectors described in the previous section, i.e. the word embedding and the character embedding, and the result is then fed into the BiLSTM layer. 
Previous taggers using DMII have had to map the information to the IFD tagset. As the tagsets of IFD and DMII are not completely compatible some information has been lost in the mapping process.
Our method allows the model to use and learn from all the information encoded in the morphological lexicon, even though it uses a tagset slightly different from our training data.



\subsection{Stepwise Tagging Model}
\label{sec:stepwiseModel}

When employing a fine-grained tagset with mnemonic encoding, the model does not place different significance on two tags when they differ in lexical category, on one hand, or share a lexical category but differ in morphological features, on the other. 
A human, however, would consider the former a more significant error than the latter. A PoS tagger is especially prone to such errors when the tagset is large and the amount of training data is insufficient to detect all the subtle differences between labels, as sometimes is the case for under-resourced or domain-specific languages.


To place a higher emphasis on assigning the correct lexical category, we devise a two-step process. First, we simplify the tagset from 565 to 10 tags by using only the first letter of the fine-grained tag mnemonic, i.e. the letter denoting the lexical category. We then train our model on this new coarse-grained tagset, using word and character embeddings as well as the morphological lexicon. This results in a lexical category tagger with very high accuracy, 98.97\% in our case. In the second step, the output of that tagger is embedded as a one-hot vector and concatenated to the vectors input into the BiLSTM layer of the main model. This guides the tagger to the correct lexical category and eliminates some of the errors caused by insufficient training data. 

This is a work in progress and other morphological features in the tags are promising for evolving this stepwise approach and further increasing overall accuracy. Thus, separate models for detecting gender, number and case agreement, for example, might be considered at each step.

We are not aware of other implementations of stepwise PoS tagging using BiLSTMs, but \citet{DBLP:conf/coling/HorsmannZ16} employ such a method in a slightly different setting. They use a Support Vector Machine for training, assume the coarse-grained tags are correct and then have their tagger assign the fine-grained tags based on them. In their Bidir tagger, \citet{dredze2008a} tag case separately in a second-pass, after running a general first-pass that uses the whole tagset. The second-pass tagger has access to the output of the first-pass, and is permitted to change its case and gender selections.

\begin{figure*}
\centering
\includegraphics[width=\textwidth]{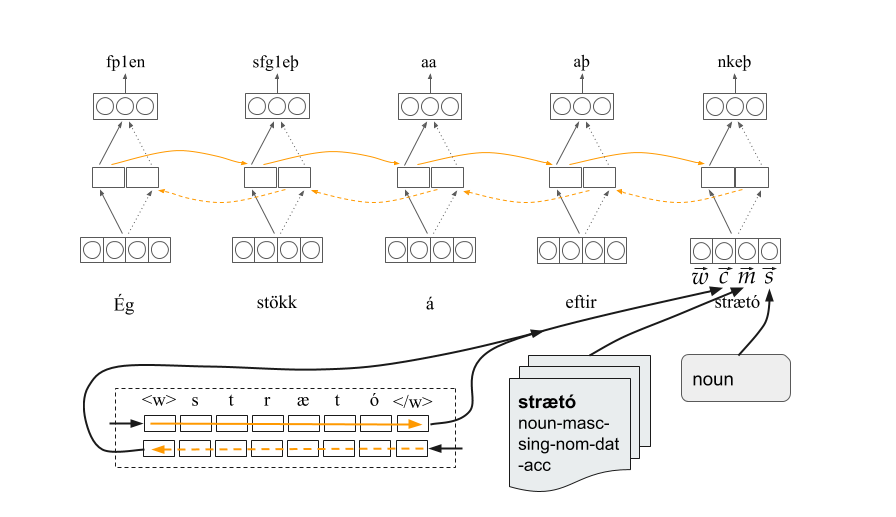}
\caption{\label{fig-model}Our full model, employing word embeddings, character embeddings, a morphological lexicon, and the output of the first-pass of the stepwise model. The hidden layer is omitted for simplicity. Figure adapted from \cite{DBLP:journals/corr/PlankSG16}.}
\end{figure*}

\section{Experiments and Results}

\subsection{Experimental Setup}
Our models were built using DyNet\footnote{The Dynamic Neural Network Toolkit, see \url{http://dynet.io}.} \citep{DBLP:journals/corr/NeubigDGMAABCCC17}. We use the same hyperparameters for all models, 
SGD training with the initial learning rate of 0.13, which decays 5\% in each epoch and runs for 30 epochs. The network has 128-dimensional embeddings for words and 20 for characters. The supplemental embeddings have 61 dimensions for the lexicon and 10 for the lexical categories. The hidden layer has 32 hidden states.\footnote{The source code for our implementation is available from \url{https://github.com/steinst/ABLTagger}}

Our experiments consist of three models:

\textbf{Baseline:} The first model uses word and character embeddings only. This corresponds to the model described in Section \ref{sec:wordCharEmbeddings}.

\textbf{DMII:} The second model adds external morphological data from DMII to the baseline model by encoding the information in n-hot vectors as described in Section \ref{sec:DMII}.

\textbf{LC:} The third and full model then adds the lexical category embeddings created by a coarse-grained tagging step described in Section \ref{sec:stepwiseModel}. This full model is shown in Figure \ref{fig-model}.


\begin{table}[t]
\begin{center}
\begin{tabular}{|l|c|c|c|}
\hline
 & \multicolumn{1}{c|}{Acc.} & \multicolumn{1}{c|}{Known (\%)} & \multicolumn{1}{c|}{Unknown (\%)}\\ 
 \hline
Baseline               & 93.25 & 95.19 & \bf 66.84\\
+ DMII          & 94.84 & 95.17 & 54.61\\
+ LC     & \bf 95.15 & \bf 95.48 & 54.06\\

\hline
\end{tabular}
\end{center}
\caption{Accuracy of the three models trained and tested on IFD. Note that when DMII is employed the number of unknown words falls almost 90\%, from 4,036 to 476 out of an average total of 58,977 words in the splits.}
\label{tab:results}
\end{table}

\subsection{Part-of-Speech Tagging Results}

The test results for all three models are shown in Table~\ref{tab:results}, with the full model reaching 95.15\% accuracy after 30 epochs. 
The baseline model (93.25\%) already gets close to state-of-the-art results and surpasses existing taggers when not using an external morphological lexicon (cf. \citealt{Loftsson2013tagginga}). 

The substantial gain achieved by using DMII confirms the advantages of using an external morphological lexicon as discussed in Section \ref{sec:DMII}. The accuracy gain is considerably higher than the corresponding gain in IceStagger (1.59 vs. 0.88 percentage points).

By employing the stepwise model discussed in Section \ref{sec:stepwiseModel} we try to guide the tagger to the highly accurate lexical category given by the coarse-grained tagger. This helps in assigning rare or ambiguous tags in the fine-grained tagset by raising the accuracy of the lexical category, resulting in a further 0.31 percentage point gain.

Note that the baseline model achieves the highest accuracy for unknown words because when adding data from DMII the unknown word ratio drops considerably (see Table \ref{tab:results}), from 6.8\% to 0.8\%. This is in line with results of previous taggers (see Section \ref{sec:data}), further reduction in unknown words is due to us using the latest version of DMII, while previous results were published in 2011. When DMII is employed the remaining unknown words are more likely to be foreign words, typos or to be irregular in some other way and therefore more difficult to tag. This explains the drop in accuracy for unknown words.

\begin{table}[t]
\begin{center}
\begin{tabular}{|l|c|c|c|c|r|}
\hline
 & \multicolumn{1}{c|}{Acc.} & \multicolumn{1}{c|}{Known} & \multicolumn{1}{c|}{Unknown} \\ \hline
TnT                        & 90.45\% & 91.82\% & 71.82\% \\
IceTagger                  & 92.73\% & 93.84\% & \bf 77.47\% \\
+ DMII                  & 93.48\% & 93.85\% & 60.50\% \\
IceStagger                 & 92.82\% & 93.97\% & 77.03\% \\
+ DMII                 & 93.70\% & 94.02\% & 61.45\% \\
+ DMII,WE           & 93.84\% & 94.15\% & 61.99\% \\
\bf Our model              & \bf 95.15\% & \bf 95.48\% & 54.06\% \\
\hline
\end{tabular}
\end{center}
\caption{Comparison to other taggers for Icelandic.}
\label{tab:taggers} 
\end{table}

\subsection{Comparison to other Taggers}

A comparison of our model to other previously published taggers for Icelandic is shown in Table~\ref{tab:taggers}. 
The results for TnT, IceTagger and IceStagger are presented in \cite{Loftsson2009ImprovingTP,loftsson_2008,Loftsson2013tagginga}, respectively.
All the reported results are fully comparable as they are based on exactly the same cross-validation split of the IFD corpus, with the exception that the TnT tagger does not employ data from DMII, and has therefore a higher ratio of unknown words.

Our model outperforms all previous taggers by a substantial margin, equaling a 21.3\% reduction in errors compared to the highest accuracy (93.84\%) obtained by IceStagger. It also has the highest accuracy for known words, i.e. those seen in the training data, including DMII. It should be noted though, that the numbers for accuracy of known and unknown words are not very well comparable between the different models, as using DMII eliminates a substantial part of unknown words, but the ones that remain tend to be more irregular, and can thus be harder to tag correctly.

\subsection{Error Analysis}
\label{sec:errors}
When comparing the most frequent kinds of errors our tagger makes to the errors of IceStagger, two differences stand out. The frequency of sng>sfg3fn and sfg3eþ>sfg1eþ are drastically reduced and are no longer among the ten most frequent kinds of errors (see Table~\ref{tab:errors}). These are verbs that are assigned infinitive mood instead of indicative mood (sn...>sf...) and third person instead of first (sfg3...>sfg1...), respectively. These kinds of errors occur when the subject is far away from the verb itself and the more frequent tag for the word form is selected instead of the correct one. This corroborates that LSTMs are better at handling long-distance dependencies \cite{DBLP:journals/corr/LinzenDG16} than other methods that have a limited context window during training.

\begin{table}[t]
\begin{center}
\begin{tabular}{|r|l|r||l|}
\hline
 & \multicolumn{2}{l||}{Our model} & \multicolumn{1}{l|}{IceStagger} \\
\hline
& \multicolumn{1}{l|}{Proposed tag} & \multicolumn{1}{l||}{Error} & \multicolumn{1}{l|}{Proposed tag} \\
\multicolumn{1}{|l|}{No.} & \multicolumn{1}{l|}{> gold tag} & \multicolumn{1}{l||}{rate} & \multicolumn{1}{l|}{> gold tag} \\
\hline
1.  & aþ>ao     & 3.28\% & aþ>ao         \\
2.  & ao>aþ     & 2.99\% & ao>aþ         \\
3.  & nveo>nveþ & 1.80\% & nveo>nveþ     \\
4.  & nveþ>nveo & 1.72\% & nveþ>nveo     \\
5.  & ao>aa     & 1.18\% & \bf sng>sfg3fn    \\
6.  & aa>ao     & 1.09\% & ao>aa         \\
7.  & nkeo>nkeþ & 0.98\% & \bf sfg3eþ>sfg1eþ \\
8.  & nheo>nhfo & 0.92\% & aa>ao         \\
9.  & nkeþ>nkeo & 0.82\% & nheo>nhen     \\
10. & ct>c      & 0.81\% & nhen>nheo     \\
\hline
\end{tabular}
\end{center}
\caption{Ten most frequent kinds of errors.}
\label{tab:errors}
\end{table}

The remaining kinds of errors in the top ten list are for the most part mistakes in case assignment. For example, prepositions are often wrongly marked as governing accusative instead of dative and vice versa (1 and 2) and there is often a confusion between prepositions and adverbs (5 and 6). The same goes for nouns (7 to 9) and, in addition, they are often assigned the wrong number, i.e. singular instead of plural (8). 
The last kind of error (10) is caused by a lack of syntactic and contextual information: a conjunction is marked as a relativizer, i.e. conjunction introducing a relative clause. 

The nearly even distributions (1+2, 3+4, 5+6, 7+9) at which these kinds of errors occur indicate that there is nothing in the training data to discern which tag to select in these instances. One way forward to try to tackle these errors is to supplement the model further, e.g. with verb sub-categorization frames.

\section{Tagging a Different Gold Standard}
In the previous sections, we have described the tagging process and compared the results to previous taggers using the same splits on IFD. We have demonstrated that our tagger achieves a significant gain in accuracy over previous taggers. Since the IFD corpus mainly contains literary work (see Section \ref{sec:data}), these texts are not necessarily characteristic of texts that have to be tagged for language technology or research purposes. This is one of the reasons why a new gold standard, MIM-GOLD, was built containing more diverse texts (see Section \ref{sec:data}). In 2015, \citet{steingrimsson-etal-2015-analysing} trained IceStagger on MIM-GOLD, but found it had many inconsistencies and errors. Since then it has been reviewed and corrected and the final version, along with 10-fold splits, was made available in 2018.

\begin{table}[t!]
\begin{center}
\begin{tabular}{|l|c|c|c|}
\hline
 & \multicolumn{1}{c|}{Acc.} & \multicolumn{1}{c|}{Known} & \multicolumn{1}{c|}{Unknown} \\ \hline
MIM-GOLD               & 94.04\% & 95.13\% & 68.34\% \\
+ IFD          & 94.17\% & 95.62\% & 68.18\% \\

\hline
\end{tabular}
\end{center}
\caption{Accuracy when training and testing on MIM-GOLD.}
\label{tab:results-mim-gold}
\end{table}

We trained our BiLSTM tagger on these splits and measured the accuracy for our full model, employing both DMII and the two-step method. We carried out two experiments. In the first, we only trained and tested on the 10-fold splits for MIM-GOLD, but in the second we added the whole IFD corpus to the training data. 
As evident from Table \ref{tab:results-mim-gold}, there is a substantial drop in accuracy compared to training and testing on IFD (see Table \ref{tab:results}). The lower accuracy may, at least partly, be due to a greater variety in texts than before and a larger proportion of unknown words in the MIM-GOLD test set compared 
to IFD \cite{steingrimsson-etal-2015-analysing}. 

\section{Conclusions and Future Work}

We have shown that BiLSTM models with combined word and character embeddings achieve state-of-the-art accuracy in PoS tagging of Icelandic texts. We have also confirmed that BiLSTMs perform well with a fine-grained tagset, such as the one used in the Icelandic corpora, IFD and MIM-GOLD. When dealing with small corpora, as often is the case with under-resourced languages, supplementing the models with external data can be highly beneficial as shown by our experiments. 

To deal with the problem of data sparsity, which is more prevalent when using fine-grained tagsets, we devised a stepwise method to guide the tagger in assigning lexical categories. This method is a work in progress -- we have pinpointed morphological features that can be independently identified with very high accuracy and are therefore promising candidates for being handled in a separate step in the tagging process. Furthermore, it could be worthwhile to pre-train word embeddings on unlabeled data, such as the Icelandic Gigaword Corpus (IGC) of 1.2 billion words
\cite{STEINGRIMSSON18.746}, e.g. employing the method described by \citet{DBLP:journals/corr/WangQSHZ15}, which is specifically adapted to BiLSTMs.

The error analysis in Section \ref{sec:errors} suggests that information on case governance is critical in reducing the most common errors the tagger makes. This could be external data on case governance of verbs and prepositions, or data derived from a method akin to the stepwise method that better discerns this information from the training data.

The final version of a new gold standard, MIM-GOLD, has recently been released and has not been used for training a PoS tagger for Icelandic before. IFD is heavily biased towards literary fiction but MIM-GOLD is a more balanced mix of different text genres and is thus more diverse. The lower accuracy for MIM-GOLD should thus not have been surprising, even though it has more data than IFD. Comparison of error analysis for both gold standards should reveal if there are other factors at play. We suggest that further work on developing PoS taggers for Icelandic texts focuses on this new gold standard.

\bibliography{acl2019}

\begin{thebibliography}{24}
\expandafter\ifx\csname natexlab\endcsname\relax\def\natexlab#1{#1}\fi

\bibitem[{Bjarnadóttir(2012)}]{Bjarnadottir2012}
Kristín Bjarnadóttir. 2012.
\newblock \href {https://www.aflat.org/files/saltmil8-aflat2012.pdf\#page=25}
  {{The Database of Modern {I}celandic Inflection}}.
\newblock In \emph{{{LREC} 2012 {Proceedings: {P}roceedings of Language
  Technology for Normalization of Less-Resourced Languages, SaLTMiL 8 –
  AfLaT}}}.

\bibitem[{Dos~Santos and Zadrozny(2014)}]{DosSantos:2014:LCR:3044805.3045095}
C\'{\i}cero~Nogueira Dos~Santos and Bianca Zadrozny. 2014.
\newblock \href {http://dl.acm.org/citation.cfm?id=3044805.3045095} {{Learning
  Character-level Representations for Part-of-Speech Tagging}}.
\newblock In \emph{Proceedings of the $31^{st}$ International Conference on
  Machine Learning -- Volume 32}, ICML '14, Beijing, China.

\bibitem[{Dredze and Wallenberg(2008)}]{dredze2008a}
Mark Dredze and Joel Wallenberg. 2008.
\newblock Further results and analysis of {Icelandic} part of speech tagging.
\newblock Technical report, Department of Computer and Information Science,
  University of Pennsylvania.

\bibitem[{Graves and Schmidhuber(2005)}]{Graves:2005:SIF:1120509.1120527}
Alex Graves and J\"{u}rgen Schmidhuber. 2005.
\newblock \href {https://doi.org/10.1016/j.neunet.2005.06.042} {{Framewise
  Phoneme Classification with Bidirectional LSTM and Other Neural Network
  Architectures}}.
\newblock \emph{Neural Networks}, 18(5-6):602--610.

\bibitem[{Haji{\v{c}}(2000)}]{biblio:HaMorphologicalTagging2000}
Jan Haji{\v{c}}. 2000.
\newblock \href {https://dl.acm.org/citation.cfm?id=974318} {Morphological
  tagging: Data vs. dictionaries}.
\newblock In \emph{$6^{th}$ {ANLP} Conference / $1^{st}$ {NAACL} Meeting.
  Proceedings}, Seattle, Washington.

\bibitem[{Helgad{\'o}ttir et~al.(2012)Helgad{\'o}ttir, Svavarsd{\'o}ttir,
  R{\"o}gnvaldsson, Bjarnad{\'o}ttir, and Loftsson}]{Helgadttir2012TheTI}
Sigr{\'u}n Helgad{\'o}ttir, {\'A}sta Svavarsd{\'o}ttir, Eir{\'i}kur
  R{\"o}gnvaldsson, Krist{\'i}n Bjarnad{\'o}ttir, and Hrafn Loftsson. 2012.
\newblock \href {https://www.aflat.org/files/saltmil8-aflat2012.pdf\#page=79}
  {{The Tagged Icelandic Corpus (MÍM)}}.
\newblock In \emph{Proceedings of SaLTMiL-AfLaT Workshop on Language technology
  for normalisation of less-resourced languages}, LREC 2012, Istanbul, Turkey.

\bibitem[{Helgadóttir(2005)}]{helgadottir2005}
Sigrún Helgadóttir. 2005.
\newblock Testing data-driven learning algorithms for {PoS} tagging of
  {Icelandic}.
\newblock In H.~Holmboe, editor, \emph{Nordisk Sprogteknologi 2004}. Museum
  Tusculanums Forlag, Copenhagen.

\bibitem[{Hochreiter and
  Schmidhuber(1997)}]{Hochreiter:1997:LSM:1246443.1246450}
Sepp Hochreiter and J\"{u}rgen Schmidhuber. 1997.
\newblock \href {https://doi.org/10.1162/neco.1997.9.8.1735} {Long short-term
  memory}.
\newblock \emph{Neural Computation}, 9(8):1735--1780.

\bibitem[{Horsmann and Zesch(2016)}]{DBLP:conf/coling/HorsmannZ16}
Tobias Horsmann and Torsten Zesch. 2016.
\newblock \href {http://aclweb.org/anthology/C/C16/C16-1032.pdf} {Assigning
  fine-grained {PoS} tags based on high-precision coarse-grained tagging}.
\newblock In \emph{Proceedings of {COLING} 2016, $26^{th}$ International
  Conference on Computational Linguistics}, Osaka, Japan.

\bibitem[{Horsmann and Zesch(2017)}]{D17-1076}
Tobias Horsmann and Torsten Zesch. 2017.
\newblock \href {https://doi.org/10.18653/v1/D17-1076} {Do {LSTM}s really work
  so well for {PoS} tagging? -- a replication study}.
\newblock In \emph{Proceedings of the 2017 Conference on Empirical Methods in
  Natural Language Processing}, Copenhagen, Denmark.

\bibitem[{Ling et~al.(2015)Ling, Dyer, Black, Trancoso, Fermandez, Amir,
  Marujo, and Luis}]{DBLP:journals/corr/LingLMAADBT15}
Wang Ling, Chris Dyer, Alan~W Black, Isabel Trancoso, Ramon Fermandez, Silvio
  Amir, Luis Marujo, and Tiago Luis. 2015.
\newblock \href {https://doi.org/10.18653/v1/D15-1176} {{Finding Function in
  Form: Compositional Character Models for Open Vocabulary Word
  Representation}}.
\newblock In \emph{Proceedings of the 2015 Conference on Empirical Methods in
  Natural Language Processing}, Lisbon, Portugal.

\bibitem[{Linzen et~al.(2016)Linzen, Dupoux, and
  Goldberg}]{DBLP:journals/corr/LinzenDG16}
Tal Linzen, Emmanuel Dupoux, and Yoav Goldberg. 2016.
\newblock \href {http://aclweb.org/anthology/Q16-1037} {{Assessing the Ability
  of LSTMs to Learn Syntax-Sensitive Dependencies}}.
\newblock \emph{Transactions of the Association for Computational Linguistics},
  4:521--535.

\bibitem[{Loftsson(2008)}]{loftsson_2008}
Hrafn Loftsson. 2008.
\newblock \href {https://doi.org/10.1017/S0332586508001820} {Tagging
  {Icelandic} text: A linguistic rule-based approach}.
\newblock \emph{Nordic Journal of Linguistics}, 31(1):47–72.

\bibitem[{Loftsson et~al.(2011)Loftsson, Helgad{\'{o}}ttir, and
  R{\"{o}}gnvaldsson}]{DBLP:conf/ranlp/LoftssonHR11}
Hrafn Loftsson, Sigr{\'{u}}n Helgad{\'{o}}ttir, and Eir{\'{\i}}kur
  R{\"{o}}gnvaldsson. 2011.
\newblock \href {http://www.aclweb.org/anthology/R11-1007} {{Using a
  Morphological Database to Increase the Accuracy in POS Tagging}}.
\newblock In \emph{Recent Advances in Natural Language Processing}, RANLP 2011,
  Hissar, Bulgaria.

\bibitem[{Loftsson et~al.(2009)Loftsson, Kramarczyk, Helgad{\'o}ttir, and
  R{\"o}gnvaldsson}]{Loftsson2009ImprovingTP}
Hrafn Loftsson, Ida Kramarczyk, Sigr{\'u}n Helgad{\'o}ttir, and Eir{\'i}kur
  R{\"o}gnvaldsson. 2009.
\newblock \href {https://dspace.ut.ee/handle/10062/9736} {Improving the {PoS}
  tagging accuracy of {Icelandic} text}.
\newblock In \emph{Proceedings of the $17^{th}$ Nordic Conference on
  Computational Linguistics}, NODALIDA 2009, Odense, Denmark.

\bibitem[{Loftsson and {\"O}stling(2013)}]{Loftsson2013tagginga}
Hrafn Loftsson and Robert {\"O}stling. 2013.
\newblock \href {https://aclanthology.info/papers/W13-5613/w13-5613} {{Tagging
  a Morphologically Complex Language Using an Averaged Perceptron Tagger: The
  Case of {I}celandic}}.
\newblock In \emph{Proceedings of the $19^{th}$ Nordic Conference on
  Computational Linguistics}, NODALIDA 2013, Oslo, Norway.

\bibitem[{Loftsson et~al.(2010)Loftsson, Yngvason, Helgadóttir, and
  Rögnvaldsson}]{Loftsson2010DevelopingAP}
Hrafn Loftsson, Jökull~H. Yngvason, Sigrún Helgadóttir, and Eiríkur
  Rögnvaldsson. 2010.
\newblock {Developing a PoS-tagged corpus using existing tools}.
\newblock In \emph{Proceedings of 7th SaLTMiL Workshop on Creation and Use of
  Basic Lexical Resources for Less-Resourced Languages}, LREC 2010, Valetta,
  Malta.

\bibitem[{Neubig et~al.(2017)Neubig, Dyer, Goldberg, Matthews, Ammar,
  Anastasopoulos, Ballesteros, Chiang, Clothiaux, Cohn, Duh, Faruqui, Gan,
  Garrette, Ji, Kong, Kuncoro, Kumar, Malaviya, Michel, Oda, Richardson,
  Saphra, Swayamdipta, and Yin}]{DBLP:journals/corr/NeubigDGMAABCCC17}
Graham Neubig, Chris Dyer, Yoav Goldberg, Austin Matthews, Waleed Ammar,
  Antonios Anastasopoulos, Miguel Ballesteros, David Chiang, Daniel Clothiaux,
  Trevor Cohn, Kevin Duh, Manaal Faruqui, Cynthia Gan, Dan Garrette, Yangfeng
  Ji, Lingpeng Kong, Adhiguna Kuncoro, Gaurav Kumar, Chaitanya Malaviya, Paul
  Michel, Yusuke Oda, Matthew Richardson, Naomi Saphra, Swabha Swayamdipta, and
  Pengcheng Yin. 2017.
\newblock \href {http://arxiv.org/abs/arXiv:1701.03980} {{DyNet}: The dynamic
  neural network toolkit}.
\newblock \emph{CoRR}, abs/1701.03980.

\bibitem[{Pind et~al.(1991)Pind, Magn{\'u}sson, and
  Briem}]{magnusson1991islensk}
Jörgen Pind, Friðrik Magn{\'u}sson, and Stefán Briem. 1991.
\newblock \emph{{\'I}slensk or{\dh}t{\'i}{\dh}nib{\'o}k [{T}he {I}celandic
  {F}requency {D}ictionary]}.
\newblock The Institute of Lexicography, University of Iceland, Reykjavik,
  Iceland.

\bibitem[{Plank et~al.(2016)Plank, S{\o}gaard, and
  Goldberg}]{DBLP:journals/corr/PlankSG16}
Barbara Plank, Anders S{\o}gaard, and Yoav Goldberg. 2016.
\newblock \href {https://doi.org/10.18653/v1/P16-2067} {{Multilingual
  Part-of-Speech Tagging with Bidirectional Long Short-Term Memory Models and
  Auxiliary Loss}}.
\newblock In \emph{Proceedings of the $54^{th}$ Annual Meeting of the
  Association for Computational Linguistics}, Berlin, Germany.

\bibitem[{Sagot and Mart{\'i}nez~Alonso(2017)}]{W17-6304}
Beno{\^i}t Sagot and H{\'e}ctor Mart{\'i}nez~Alonso. 2017.
\newblock \href {http://aclweb.org/anthology/W17-6304} {{Improving neural
  tagging with lexical information}}.
\newblock In \emph{Proceedings of the $15^{th}$ International Conference on
  Parsing Technologies}, Pisa, Italy.

\bibitem[{Steingr{\'\i}msson et~al.(2015)Steingr{\'\i}msson, Helgad{\'o}ttir,
  and R{\"o}gnvaldsson}]{steingrimsson-etal-2015-analysing}
Stein{\th}{\'o}r Steingr{\'\i}msson, Sigr{\'u}n Helgad{\'o}ttir, and
  Eir{\'\i}kur R{\"o}gnvaldsson. 2015.
\newblock \href {https://www.aclweb.org/anthology/W15-1838} {{Analysing
  Inconsistencies and Errors in {P}o{S} Tagging in two {I}celandic Gold
  Standards}}.
\newblock In \emph{Proceedings of the $20^{th}$ Nordic Conference of
  Computational Linguistics}, NODALIDA 2015, Vilnius, Lithuania.

\bibitem[{Steingrímsson et~al.(2018)Steingrímsson, Helgadóttir,
  Rögnvaldsson, Barkarson, and Guðnason}]{STEINGRIMSSON18.746}
Steinþór Steingrímsson, Sigrún Helgadóttir, Eiríkur Rögnvaldsson,
  Starkaður Barkarson, and Jón Guðnason. 2018.
\newblock \href {http://www.lrec-conf.org/proceedings/lrec2018/pdf/746.pdf}
  {{Risamálheild: A Very Large Icelandic Text Corpus}}.
\newblock In \emph{Proceedings of the Eleventh International Conference on
  Language Resources and Evaluation}, LREC 2018, Miyazaki, Japan.

\bibitem[{Wang et~al.(2015)Wang, Qian, Soong, He, and
  Zhao}]{DBLP:journals/corr/WangQSHZ15}
Peilu Wang, Yao Qian, Frank~K. Soong, Lei He, and Hai Zhao. 2015.
\newblock \href {https://arxiv.org/pdf/1510.06168.pdf} {{Part-of-Speech Tagging
  with Bidirectional Long Short-Term Memory Recurrent Neural Network}}.
\newblock \emph{CoRR}, abs/1510.06168.

\end{thebibliography}
\bibliographystyle{acl_natbib}

\appendix

\end{document}